\documentclass[hyphens]{article}
\usepackage[final]{neurips_2025}

\usepackage[utf8]{inputenc}
\usepackage[T1]{fontenc}
\usepackage{hyperref}
\hypersetup{
colorlinks=true,
linkcolor=red,
citecolor=cyan,
filecolor=magenta,
urlcolor=magenta,
}
\usepackage{url}
\usepackage{xurl}
\usepackage{booktabs}
\usepackage{amsfonts}
\usepackage{nicefrac}
\usepackage{microtype}
\usepackage[table,dvipsnames]{xcolor}
\usepackage{amsmath}
\usepackage{amssymb}
\usepackage{cleveref}
\usepackage{xspace}
\usepackage{enumitem}
\usepackage{textcomp}
\usepackage{stfloats}
\usepackage{verbatim}
\usepackage{wrapfig}
\usepackage{graphicx}
\usepackage{float}
\usepackage[numbers,sort&compress]{natbib}
\usepackage{tikz}
\usepackage{algorithm}
\usepackage{algpseudocode}
\usepackage{makecell}
\usepackage{multicol,multirow}
\usepackage{threeparttable}
\usepackage{tablefootnote}
\usepackage{pgfplots}
\usepackage[labelfont=bf]{caption}
\usepackage{tabularx}
\usepackage{array}
\usepackage{ragged2e}
\usepackage[section]{placeins}
\pgfplotsset{compat=1.18}

\AtBeginEnvironment{thebibliography}{\raggedright\sloppy}
\hfuzz=\maxdimen
\vfuzz=\maxdimen

\usepackage{fancyhdr}
\renewcommand{\headwidth}{\textwidth}
\renewcommand{\headrulewidth}{0.5pt}
\renewcommand{\headrule}{\vspace{2pt}\hbox to\headwidth{\color{black}\leaders\hrule height \headrulewidth\hfill}}
\newcolumntype{Y}{>{\RaggedRight\arraybackslash}X}
\DeclareRobustCommand{\xcottoken}[1]{\mbox{\texttt{\footnotesize #1}}}
\newcommand{\safeincludegraphics}[2][]{%
\IfFileExists{#2}{\includegraphics[#1]{#2}}{%
\fbox{\parbox[c][0.18\textheight][c]{0.90\linewidth}{\centering
Missing figure file:\\[2mm]\texttt{\detokenize{#2}}}}%
}%
}

\title{XCoT-VLA: Executable Chain-of-Thought for Vision-Language-Action Driving}
\author{
\colorbox{white}{\includegraphics[height=0.8em]{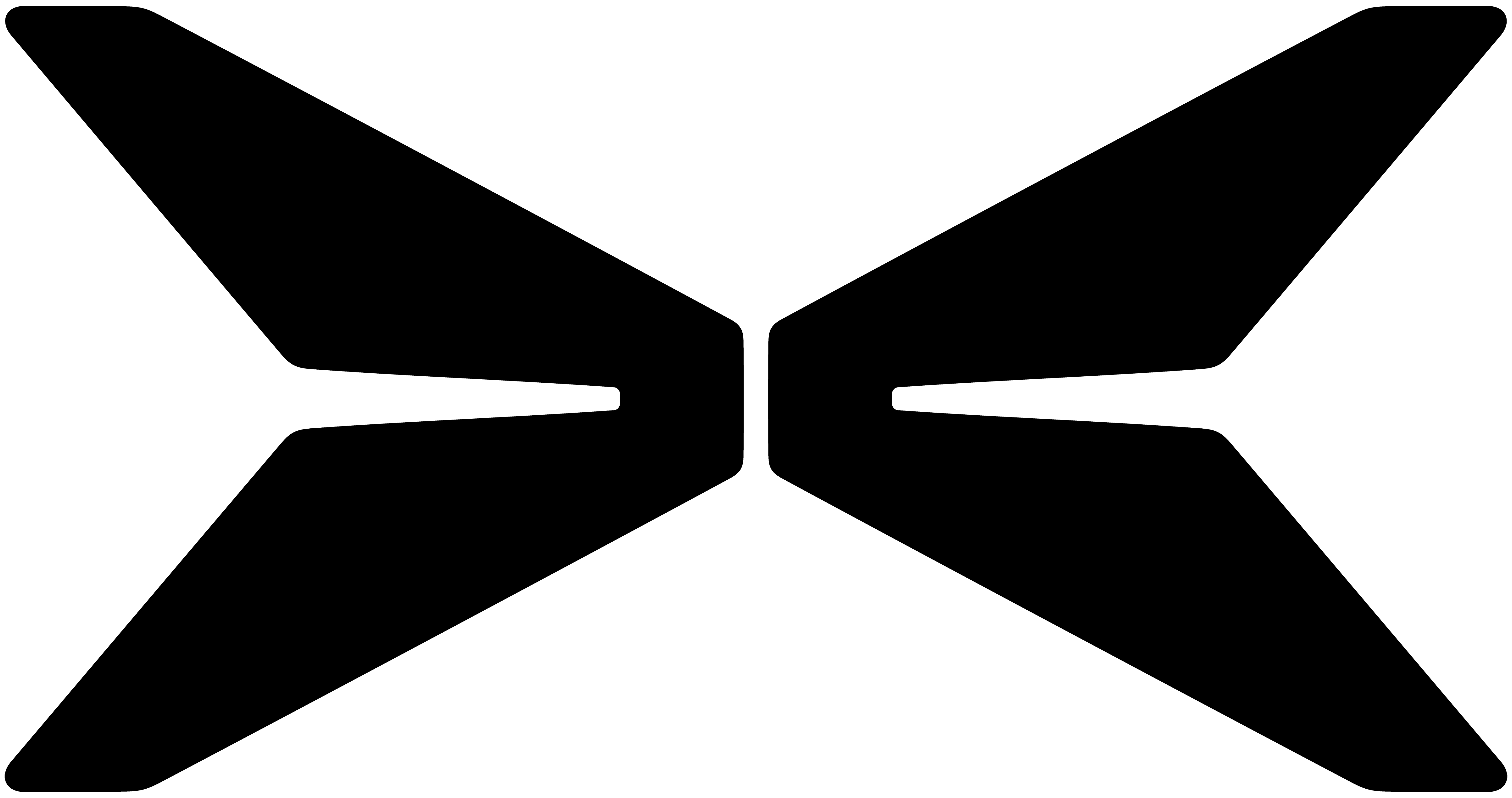}}
Foundation Model Team, XPeng Inc.\\
}

\vspace{-0.5em}
\begin{document}
\maketitle
\vspace{-0.8em}
\begin{abstract}
Vision-Language-Action (VLA) models can connect scene understanding, semantic reasoning, and trajectory generation for autonomous driving. However, verbose natural-language Chain-of-Thought (CoT) is poorly suited to real-time control because it is open-ended, costly to decode, and difficult to optimize as an action-facing representation. We propose \textbf{XCoT-VLA}, which replaces descriptive rationales with compact executable CoT tokens learned from automatically constructed Reason--Action supervision. Logged trajectories provide action evidence, while scene context supplies causal semantics. The predicted XCoT sequence remains in context and conditions fixed trajectory queries through shared multimodal self-attention. Deterministic token-function routing applies the Reason FFN to XCoT tokens and the Control FFN to trajectory queries for flow-matching trajectory generation. We further introduce XCoT Policy Optimization (XCPO) as an optional refinement extension in the same executable token space. XCoT-VLA reduces longitudinal ADE from 1.645 to 1.323 on a general-distribution set and lateral FDE from 1.616 to 0.648 in lane-change scenarios. By representing driving-oriented reasoning with only 2--6 executable XCoT tokens, our method substantially reduces autoregressive reasoning overhead and remains within the real-time planning budget. These results demonstrate that driving-oriented reasoning can be compact, executable, and directly connected to trajectory generation.
\end{abstract}
\vspace{-0.8em}

\par\vspace{0.25em}
\noindent\begin{minipage}{\textwidth}
\centering
\captionsetup{font=small,skip=2pt}
\safeincludegraphics[
width=1.0\textwidth,
height=0.8\textheight,
keepaspectratio
]{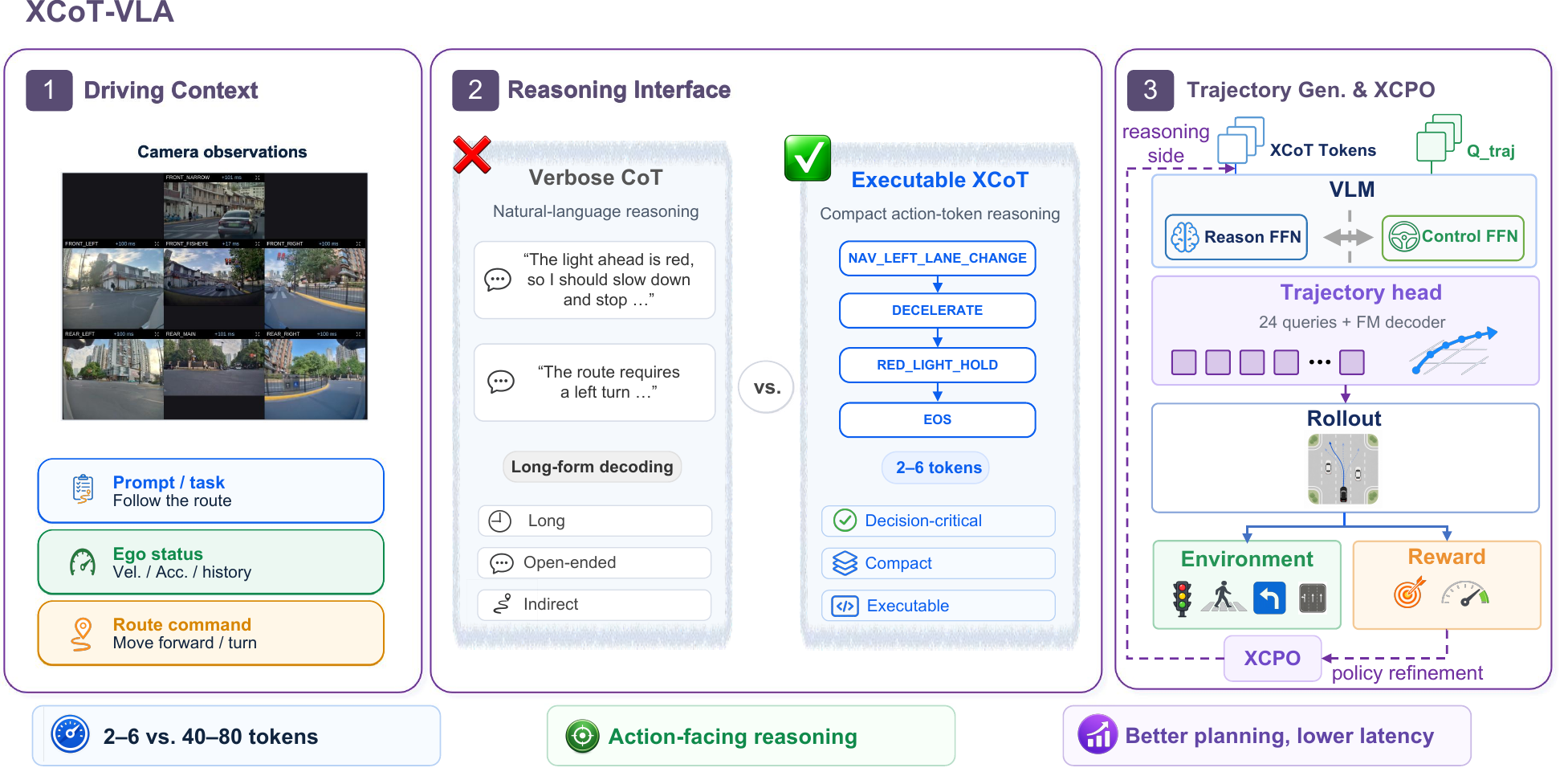}
\captionof{figure}{
Overview of XCoT-VLA.
Given multimodal driving context, XCoT-VLA replaces verbose natural-language CoT with 2--6 compact executable action tokens that capture decision-critical driving intents and condition flow-matching trajectory generation.
The same XCoT space also supports XCPO for reward-driven reasoning-policy refinement with a frozen execution stack.
}
\label{fig:arch_comparison}
\end{minipage}
\vspace{-0.35em}
\section{Introduction}

Vision-language models (VLMs) have made substantial progress in visual understanding, instruction following, and multimodal reasoning~\cite{alayrac2022flamingo,li2023blip2,liu2023llava}.
Chain-of-Thought (CoT) further enables large language and vision-language models to perform explicit intermediate reasoning before producing final predictions~\cite{wei2022chain,wang2023selfconsistency}.
Building on these advances, Vision-Language-Action (VLA) models extend semantic understanding and reasoning to physical control.
They have been explored in both robotics~\cite{zitkovich2023rt2,kim2024openvla,black2025pi0} and autonomous driving~\cite{drivevlm,shao2024lmdrive,opendrivevla,autovla}, where high-level semantics must ultimately be translated into executable actions.

In this work, we focus on autonomous driving, where reasoning and trajectory generation must operate continuously under strict real-time constraints.
Recent driving VLM/VLA methods increasingly introduce language-based reasoning or CoT-style intermediate representations to improve semantic understanding and decision making~\cite{sima2024drivelm,nie2025reason2drive,alphadrive,autovla}.
However, verbose natural-language CoT is not necessarily an effective interface for control.
Free-form rationales may contain information irrelevant to the immediate decision, their linguistic structure is only implicitly connected to executable motion, and autoregressive token-by-token decoding introduces additional latency~\cite{leviathan2023fast}.

The key question is therefore not whether driving VLA models should reason, but \emph{what form of reasoning should be exposed to the trajectory generator}.
We argue that such reasoning should be \emph{decision-critical}, retaining only semantics that affect driving behavior; \emph{compact}, requiring limited autoregressive decoding; and \emph{executable}, directly conditioning continuous trajectory generation.

To this end, we propose \textbf{XCoT-VLA}, an autonomous-driving VLA framework based on \emph{Executable Chain-of-Thought} (XCoT), as illustrated in \autoref{fig:arch_comparison}.
Instead of generating unrestricted natural-language rationales, XCoT represents reasoning as a short sequence of executable semantic-action tokens.
Tokens such as \xcottoken{LEFT\_TURN\_PREPARE}, \xcottoken{DECELERATE}, and \xcottoken{RED\_LIGHT\_HOLD} directly encode decision-critical driving intents and their control-relevant semantics.
In our driving setting, only \textbf{2--6 XCoT tokens} are typically required to represent the reasoning needed for trajectory generation, substantially reducing autoregressive reasoning overhead while satisfying the real-time planning budget.

XCoT-VLA learns this representation from structured Reason--Action supervision.
Logged trajectories provide action evidence, while scene context provides causal semantics such as navigation intent, traffic rules, road structure, and surrounding interactions.
During inference, the predicted XCoT sequence conditions fixed trajectory queries through shared multimodal self-attention.
Deterministic token-function routing processes XCoT tokens with the Reason FFN and trajectory queries with the Control FFN for flow-matching trajectory generation~\cite{lipman2023flow}.
The same executable token space can further support XCoT Policy Optimization (XCPO), which refines the reasoning policy using trajectory-level rewards while keeping the action-side executor fixed.

Our contributions are summarized as follows:
\begin{itemize}
\item \textbf{Executable CoT for real-time driving VLA.}
We introduce XCoT, a compact action-facing reasoning representation that replaces verbose free-form CoT with executable semantic-action tokens, reducing autoregressive reasoning overhead under real-time driving constraints.

\item \textbf{Decoupled reasoning--action computation.}
We construct XCoT supervision from logged action evidence and causal scene semantics, and use shared multimodal attention with separate Reason and Control FFN branches to connect reasoning with continuous trajectory generation.

\item \textbf{Deployment-oriented evaluation.}
Experiments on general-distribution planning, lane-changing scenarios, inference latency, training stability, and qualitative trajectory analysis show that XCoT improves decision-centric planning while remaining compatible with real-time inference.
\end{itemize}

Although evaluated on autonomous driving, the underlying principle of XCoT can extend to other latency-sensitive VLA systems: reasoning exposed to the action generator should be decision-critical, compact, and directly executable rather than verbose free-form language.

\section{Related Work}

\subsection{Vision-Language-Action Models}

A key question in Vision-Language-Action (VLA) models is how high-level multimodal representations are translated into actions. 
In robotics, RT-2 and OpenVLA formulate actions as discrete tokens that can be predicted within a language-model backbone~\cite{zitkovich2023rt2,kim2024openvla}, while $\pi_0$ adopts flow-based continuous action generation for general robot control~\cite{black2025pi0}. 
Related VLM/VLA systems in autonomous driving connect visual observations and language representations with driving decisions or continuous trajectory generation~\cite{drivegpt4,omnidrive,opendrivevla}.
XCoT-VLA differs in the role of discrete tokens. 
XCoT tokens are not final low-level actions; instead, they serve as an intermediate semantic-action interface between multimodal reasoning and continuous control. 
They explicitly encode decision-critical intent while directly conditioning downstream trajectory generation.

\subsection{Reasoning Interfaces for Action Generation}

Language-based intermediate reasoning has been increasingly explored for decision making. 
DriveLM organizes driving knowledge through structured question-answer reasoning, LingoQA focuses on language-based driving understanding, and Reason2Drive introduces chain-based reasoning for driving decisions~\cite{sima2024drivelm,marcu2024lingoqa,nie2025reason2drive}. 
Recent methods further integrate reasoning with planning or policy optimization~\cite{alphadrive,autovla}.
Most of these approaches represent intermediate reasoning using natural language or structured textual forms. 
XCoT instead focuses on the reasoning-to-action interface: it compresses control-relevant semantics into a short sequence of executable semantic-action tokens. 
This representation directly conditions action generation while avoiding the unnecessary autoregressive overhead of verbose free-form reasoning.

\subsection{Structured Action Spaces and Policy Optimization}

Structured action representations provide an alternative to directly exploring low-level continuous action spaces. 
Options introduce temporally abstract decisions~\cite{sutton1999options}, while latent-action methods learn compact intermediate representations for reinforcement learning and control~\cite{zhao2019larl,zhou2021plas}. 
Recent work further learns latent action spaces from language or video for downstream decision making~\cite{jia2025cola,ye2025lapa}.
Reasoning models have also begun to optimize intermediate decision processes using reinforcement learning~\cite{shao2024deepseekmath,deepseekr1}. 
XCPO follows this structured-optimization perspective but operates in an explicit and interpretable XCoT space rather than an unconstrained language or latent space. 
It evaluates sampled XCoT sequences through their induced trajectories and optimizes the reasoning policy while keeping the execution stack fixed.

\section{Method}

We model autonomous driving as a Partially Observable Markov Decision Process (POMDP)~\cite{kaelbling1998planning,hubmann2017decision}. At each decision step, the ego vehicle receives a multimodal observation $o_t$ containing the current visual context, the task prompt, ego status (including velocity, acceleration, and historical trajectory), and the high-level navigation command. The reasoning pathway first autoregressively predicts an XCoT action-token sequence $z_t$. Conditioned on both $o_t$ and $z_t$, the flow-matching trajectory head predicts a future motion sequence
\[
u_t
=
\{(a_{\mathrm{lon},h},\Delta\psi_h)\}_{h=1}^{H}
\in\mathbb{R}^{H\times 2},
\]
where $a_{\mathrm{lon},h}$ and $\Delta\psi_h$ denote longitudinal acceleration and yaw change at future step $h$, respectively. The trajectory head then temporally integrates this sequence, using the current ego state as initialization, to obtain the ego-centric coordinate trajectory
\[
\tau_t
=
\mathcal{I}_{\mathrm{traj}}(u_t;s_t)
=
\{(x_{t,h},y_{t,h})\}_{h=1}^{H}
\in\mathbb{R}^{H\times 2}.
\]
In our implementation, $H=24$. Here, \emph{XCoT policy rollout} refers only to autoregressive token sampling during XCPO; motion decoding and temporal integration are deterministic model operations. The goal of XCoT-VLA is to expose a reasoning signal that is useful for trajectory generation rather than to produce a verbose natural-language explanation.

The method is organized around one central supervision chain:
\[
\begin{aligned}
&\text{logged trajectory + scene context}
\rightarrow \text{Reason--Action supervision} \\
&\rightarrow \text{executable XCoT action-token sequence}
\rightarrow \text{trajectory decoding}.
\end{aligned}
\]
In other words, XCoT tokens are not introduced as arbitrary labels. They are constructed offline from logged driving through a data-construction pipeline that links observed motion with its causal scene semantics. This section first presents the XCoT training-data construction pipeline and the resulting executable token space, and then describes model training and optional policy refinement.

\paragraph{Notation.}
At decision step $t$, $o_t$ and $s_t$ denote the multimodal observation and current ego state, respectively. The executable XCoT sequence is
$z_t=(z_{t,1},\dots,z_{t,M_t})$, with
$z_{t,m}\in\mathcal{V}_{\mathrm{XCoT}}$.
We use $h$ to index future motion steps and $\alpha$ for flow-matching time. Offline XCoT supervision labels carry the superscript $*$; hats denote Stage-I model predictions when they are contrasted with their supervision targets.

\subsection{XCoT Training Data Construction}
\label{sec:reason_action}

\begin{figure}[!ht]
    \centering
    \safeincludegraphics[
        width=1.0\textwidth,
        height=0.80\textheight,
        keepaspectratio
    ]{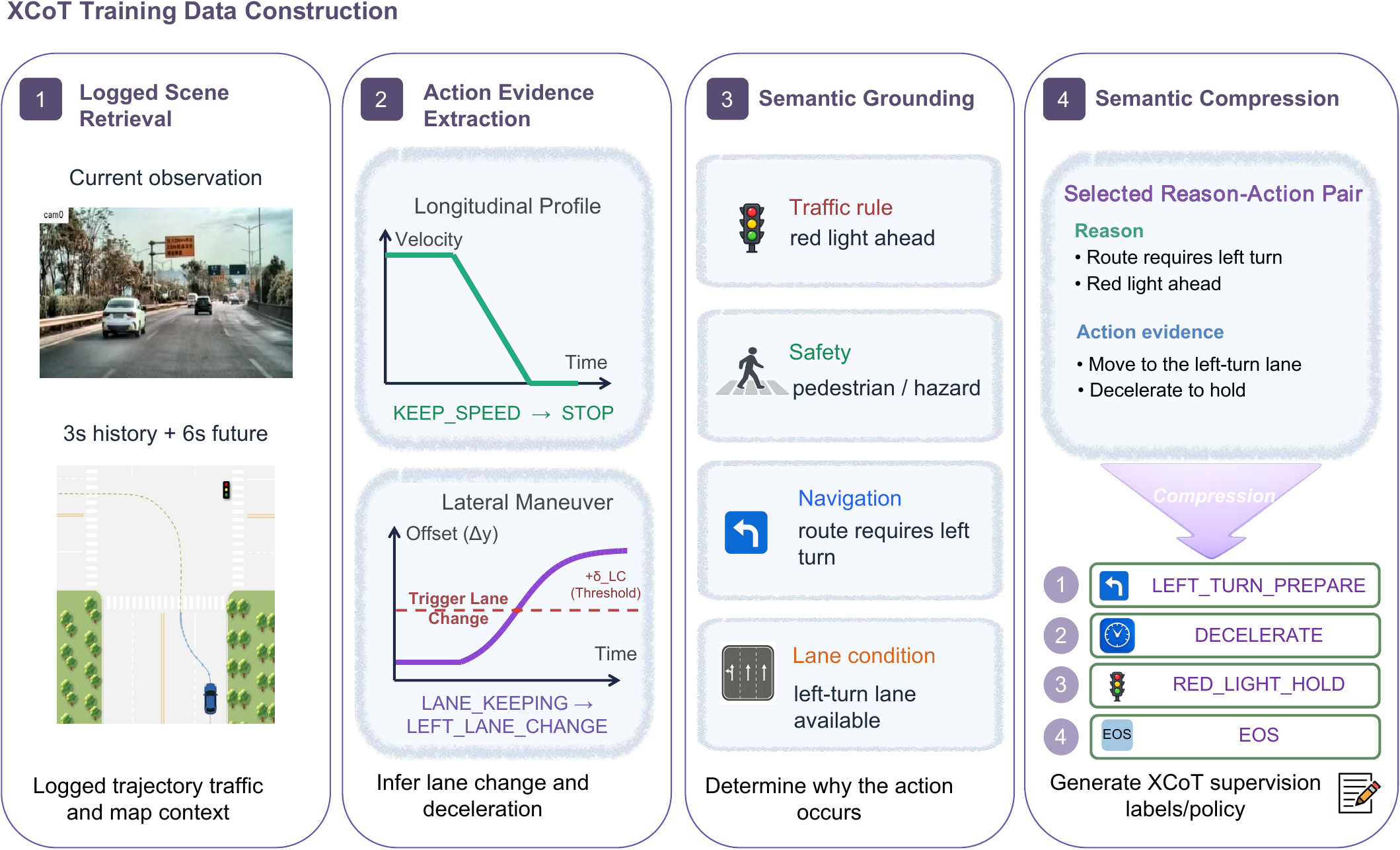}
    \caption{Offline XCoT training-data construction pipeline. The logged future trajectory provides longitudinal and lateral action evidence, while scene context supplies navigation-, rule-, interaction-, and safety-related causes. The selected Reason--Action pair is then mapped to a canonical XCoT sequence.}
    \label{fig:reason_action_construct}
\end{figure}

We formulate label construction as an offline data-construction pipeline that maps each logged sample $(o_t,\tau_t)$ to an executable XCoT sequence $z_t^*$. The pipeline consists of two stages: (i) Reason--Action pair construction, which extracts the observed action and grounds it in scene semantics; and (ii) semantic compression, which tokenizes the resulting Reason--Action pair into a compact canonical XCoT sequence. The complete pipeline is:
\begin{equation}
(o_t,\tau_t)
\xrightarrow{f_{\mathrm{act}}}
\mathbf{a}_t
\xrightarrow{f_{\mathrm{ground}}}
(r_t^*,\mathbf{a}_t)
\xrightarrow{g_{\mathrm{XCoT}}}
z_t^*.
\label{eq:data_construction_pipeline}
\end{equation}
The logged future trajectory is used only to construct training labels and is unavailable at inference time.
The overall construction process is illustrated in \autoref{fig:reason_action_construct}, including action-evidence extraction, semantic grounding, and XCoT semantic compression.

\subsubsection{Reason--Action Pair Construction}
This stage recovers both the observed driving behavior and the causal scene semantics that justify it.

\paragraph{Step 1: Extract action evidence from the logged trajectory.}
Given a logged trajectory segment $\tau_t$, we extract longitudinal and lateral behavior over the -3\,s--6\,s horizon:
\begin{equation}
\mathbf{a}_t
=
f_{\mathrm{act}}(\tau_t)
=
\left(\mathbf{a}^{\mathrm{lon}}_t,\mathbf{a}^{\mathrm{lat}}_t\right).
\label{eq:action_extraction}
\end{equation}
The resulting $\mathbf{a}_t$ is an offline evidence representation derived from velocity, acceleration, lane-relative displacement, and trajectory geometry. It identifies the observed behavior---such as speed maintenance, deceleration-to-hold, lane keeping, or a leftward maneuver---but does not by itself determine the driving intent.

\paragraph{Step 2: Ground the observed action in scene semantics.}
Because the same geometric motion can have different causes, we infer candidate reasons from the scene context conditioned on $\mathbf{a}_t$:
\begin{equation}
\mathcal{R}_t
=
f_{\mathrm{reason}}(o_t,\mathbf{a}_t)
=
\{r_t^{(k)}\}_{k=1}^{K}.
\label{eq:reason_candidates}
\end{equation}
The candidates cover navigation intent, road structure, traffic rules, surrounding-agent interactions, and safety constraints. Because all candidates share the same trajectory-derived action evidence $\mathbf{a}_t$, semantic grounding selects only the most consistent reason:
\begin{equation}
r_t^*
=
f_{\mathrm{ground}}(o_t,\mathbf{a}_t)
=
\operatorname*{arg\,max}_{r_t^{(k)}\in\mathcal{R}_t}
S_{\mathrm{cons}}\!\left(
r_t^{(k)}
\mid o_t,\mathbf{a}_t
\right).
\label{eq:expert_consensus}
\end{equation}
Thus, $f_{\mathrm{ground}}$ comprises candidate generation by $f_{\mathrm{reason}}$ followed by consensus selection. In practice, a small set of vision-language experts is queried with a fixed prompt over $o_t$ and $\mathbf{a}_t$. Their free-form explanations are not used directly as labels. Instead, rule templates, keyword cues, and embedding similarity map each explanation to a provisional taxonomy signature used only to align the experts. The consensus score $S_{\mathrm{cons}}$ measures agreement among these signatures and selects the causal reason. Final XCoT token composition and ordering are performed only by $g_{\mathrm{XCoT}}$ in the subsequent semantic-compression stage. Thus, language serves only as an offline source of causal supervision.

\subsubsection{XCoT Tokenization via Semantic Compression}
\label{sec:xcot_definition}
Given the Reason--Action pair constructed in the preceding stage, we tokenize it through semantic compression:
\begin{equation}
z_t^*
=
g_{\mathrm{XCoT}}(r_t^*,\mathbf{a}_t)
=
(z_{t,1}^*,\dots,z_{t,M_t}^*),
\qquad
z_{t,m}^*\in\mathcal{V}_{\mathrm{XCoT}}.
\label{eq:xcot_assignment}
\end{equation}
Here $g_{\mathrm{XCoT}}$ is a taxonomy-based semantic compression function over a fixed, interpretable vocabulary. Semantic compression maps free-form causal semantics and trajectory-derived action evidence into a short canonical sequence while preserving the primary maneuver, longitudinal adjustment, and decision-critical navigation, rule, interaction, or safety constraints. The sequence length satisfies $M_t\le M_{\max}$, with $M_{\max}=6$ in our implementation; the evaluated labels contain 2--6 action-facing XCoT tokens, followed by a separately appended EOS. The pipeline does not learn or invent new tokens; it assigns a canonical executable sequence to each logged scene.

An XCoT token is an executable semantic-action token. The unified vocabulary covers longitudinal-control intents such as \xcottoken{KEEP\_SPEED} and \xcottoken{DECELERATE}; lateral or navigation-conditioned maneuvers such as \xcottoken{LANE\_KEEPING}, \xcottoken{NAV\_LEFT\_LANE\_CHANGE}, \xcottoken{RIGHT\_TURN\_PREPARE}, and \xcottoken{RIGHT\_EXIT\_PREPARE}; and rule-, interaction-, environment-, or safety-conditioned intents such as \xcottoken{RED\_LIGHT\_HOLD}, \xcottoken{LEFT\_OVERTAKE}, \xcottoken{VISIBILITY\_CAUTION}, and \xcottoken{HAZARD\_YIELD}. All labels belong to the same prediction space and can be composed to express a primary maneuver together with longitudinal adjustments and context-specific modifiers. Representative XCoT tokens and their executable semantics are summarized in \autoref{tab:xcot_examples}.

\begin{table}[!t]
\centering
\footnotesize
\setlength{\tabcolsep}{3pt}
\renewcommand{\arraystretch}{1.20}

\caption{Representative XCoT action tokens. The vocabulary includes symmetric maneuver tokens (e.g., left/right lane changes); only representative examples are listed. All tokens remain executable semantic-action representations in the same prediction space.}
\label{tab:xcot_examples}

\begin{tabular}{
@{}
>{\raggedright\arraybackslash}m{0.36\textwidth}
>{\raggedright\arraybackslash}m{0.21\textwidth}
>{\raggedright\arraybackslash}m{0.37\textwidth}
@{}}
\toprule
\textbf{XCoT token} &
\textbf{Driving semantics} &
\textbf{Executable meaning} \\
\midrule

\mbox{\texttt{\scriptsize KEEP\_SPEED}}
& Longitudinal control
& Maintain the current speed. \\
\addlinespace[2pt]

\mbox{\texttt{\scriptsize LANE\_KEEPING}}
& Lateral control
& Maintain the current lane. \\
\addlinespace[2pt]

\mbox{\texttt{\scriptsize NAV\_LEFT\_LANE\_CHANGE}}
& Navigation lane change
& Change left according to the navigation intent. \\
\addlinespace[2pt]

\mbox{\texttt{\scriptsize EFFICIENCY\_LEFT\_LANE\_CHANGE}}
& Efficiency lane change
& Change left to improve driving efficiency. \\
\addlinespace[2pt]

\mbox{\texttt{\scriptsize RED\_LIGHT\_HOLD}}
& Traffic-rule action
& Stop or remain stopped for a red light. \\
\addlinespace[2pt]

\mbox{\texttt{\scriptsize LEFT\_TURN\_PREPARE}}
& Navigation preparation
& Prepare for a navigation-required left turn. \\
\addlinespace[2pt]

\mbox{\texttt{\scriptsize LEFT\_TURN}}
& Navigation action
& Execute the left-turn maneuver. \\
\addlinespace[2pt]

\mbox{\texttt{\scriptsize HAZARD\_YIELD}}
& Safety action
& Yield to a safety-critical road user or obstacle. \\

\bottomrule
\end{tabular}


\end{table}

A driving decision can require multiple XCoT tokens. For example, a pedestrian-aware right turn may be represented by \xcottoken{RIGHT\_TURN\_PREPARE}, \xcottoken{DECELERATE}, and \xcottoken{HAZARD\_YIELD}, while an overtaking maneuver in congestion may combine \xcottoken{NAV\_LEFT\_LANE\_CHANGE}, \xcottoken{LEFT\_OVERTAKE}, and \xcottoken{DECELERATE}. The sequence should be interpreted as a canonical composition of executable intents rather than as a dense frame-level temporal trace. The label-construction pipeline places the primary lateral or navigation maneuver first, followed by optional interaction or longitudinal-control intents and then environment-, rule-, or safety-conditioned modifiers. This deterministic schema avoids multiple equivalent permutations of the same decision. Sequences are terminated with EOS and padded only for batching.

\subsection{Decoupled Reasoning-Control Architecture}
\label{sec:dual_pathway}

\begin{figure}[!htbp]
    \centering
    \safeincludegraphics[width=0.95\textwidth]{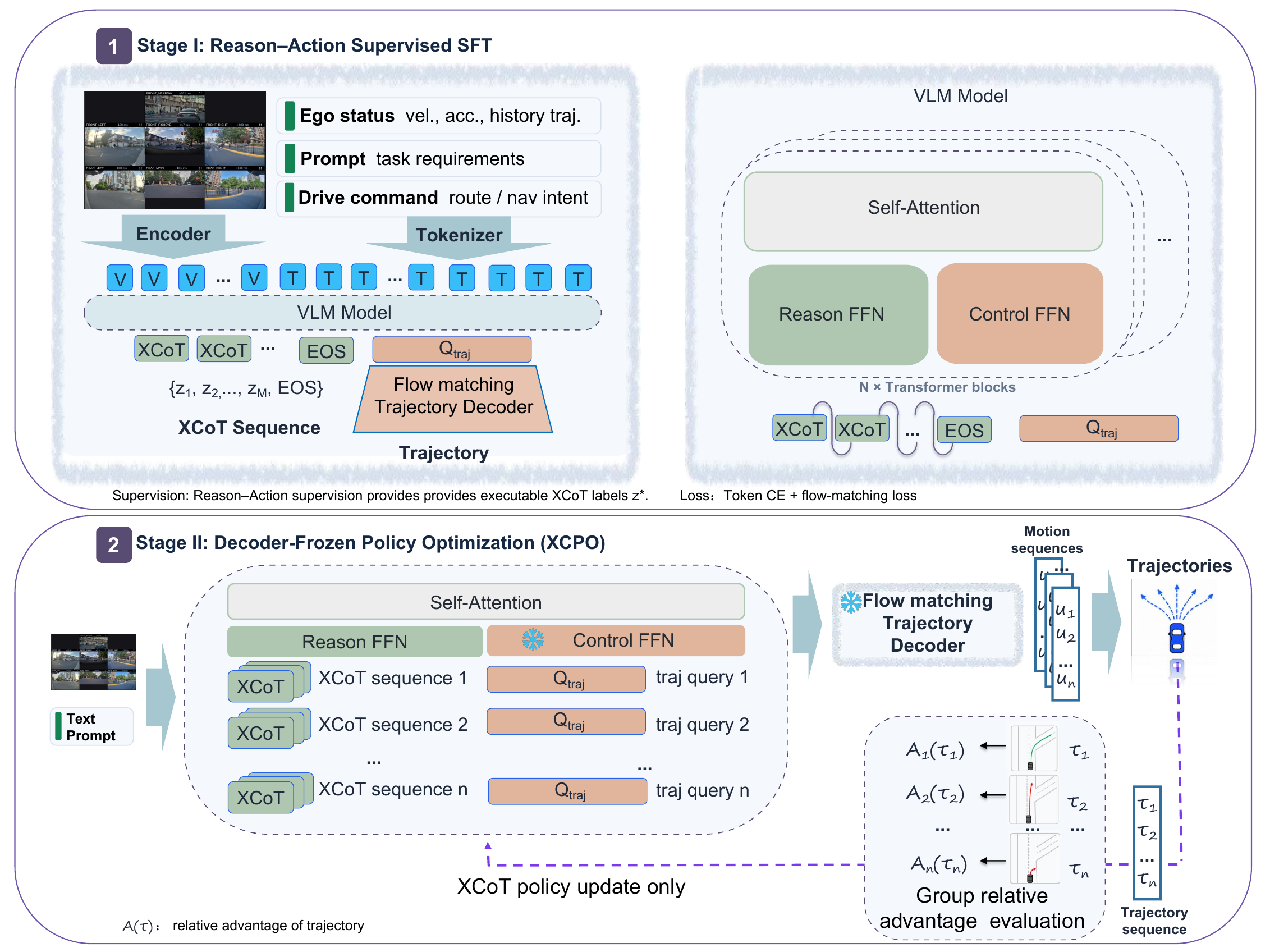}
    \caption{Overview of XCoT-VLA training and deterministic token-function routing. All valid non-trajectory tokens are processed by the Reason FFN, while only the 24 trajectory queries are processed by the Control FFN. Both branches interact through shared multimodal self-attention. In Stage II, the visual front-end, self-attention, Control FFN, and trajectory head are parameter-frozen, whereas the Reason FFN and XCoT prediction head remain trainable. Consequently, the execution stack remains fixed, but the non-trajectory representations conditioning the trajectory queries may change across policy iterations.}
    \label{fig:two_stage_training}
\end{figure}

The overall XCoT-VLA architecture, deterministic token-function routing, and two-stage training procedure are illustrated in \autoref{fig:two_stage_training}.
XCoT-VLA uses a VLA backbone with a flow-matching trajectory decoder inspired by recent action-generation models~\cite{black2025pi0}. The model contains a reasoning pathway that predicts XCoT action tokens and a control pathway that converts observation-conditioned trajectory queries into continuous future motion. We denote the fixed set of 24 learnable trajectory queries by $Q_{\mathrm{traj}}\in\mathbb{R}^{24\times d}$, with one query assigned to each future motion step. All token positions first interact through shared multimodal self-attention. After attention, deterministic token-function routing assigns every valid non-trajectory token---including visual, text, task-prompt, ego-status, drive-command, and autoregressive XCoT tokens---to the Reason FFN. Only the 24 trajectory-query positions are assigned to the Control FFN. The architecture therefore contains exactly two FFN branches: the Reason FFN and the Control FFN.

Let $H^{l-1}\in\mathbb{R}^{N\times d}$ be the token-feature matrix at the input of layer $l$. The layer first aggregates multimodal context through shared self-attention:
\begin{equation}
\tilde{H}^{l}
=
H^{l-1}
+
\mathrm{SelfAttn}\!\left(
\mathrm{Norm}(H^{l-1})
\right).
\label{eq:self_attn}
\end{equation}
We explicitly pass token-function masks to the two FFN branches. Let $m^{\mathrm{Reason}}_{t,j}$ and $m^{\mathrm{Control}}_{j}$ denote the masks for positions processed by the Reason FFN and Control FFN, respectively. Accordingly, $m^{\mathrm{Reason}}_{t,j}$ covers valid non-trajectory positions, while $m^{\mathrm{Control}}_{j}$ covers the 24 trajectory-query positions. The routed update is
\begin{equation}
\label{eq:hard_routing}
\begin{aligned}
h^l_j = \tilde{h}^l_j
&+ m^{\mathrm{Reason}}_{t,j}\,
\mathrm{FFN}_{\mathrm{Reason}}\!\left(
\mathrm{Norm}(\tilde{h}^l_j)
\right)\\
&+ m^{\mathrm{Control}}_{j}\,
\mathrm{FFN}_{\mathrm{Control}}\!\left(
\mathrm{Norm}(\tilde{h}^l_j)
\right).
\end{aligned}
\end{equation}
For each valid token position, the Reason and Control masks form a disjoint partition:
\begin{equation}
m^{\mathrm{Reason}}_{t,j}+m^{\mathrm{Control}}_{j}=1,
\qquad
m^{\mathrm{Reason}}_{t,j}m^{\mathrm{Control}}_{j}=0.
\label{eq:routing_partition}
\end{equation}
Padding positions are excluded from both masks. The Control mask is fixed over the 24 learnable trajectory-query positions, whereas the Reason mask covers all remaining valid positions and is sample-dependent because input and XCoT sequence lengths can vary. Each trajectory query represents one future motion step and is decoded into a longitudinal-acceleration and yaw-change pair, yielding $u_t\in\mathbb{R}^{24\times 2}$. The trajectory head then temporally integrates the predicted sequence to obtain $\tau_t=\mathcal{I}_{\mathrm{traj}}(u_t;s_t)$. Unlike standard MoE routing, this routing is deterministic and uses no learned router.

The Reason branch processes the complete non-trajectory context and autoregressively predicts $\hat z_t=(\hat z_{t,1},\dots,\hat z_{t,\hat M_t})$ until EOS or $M_{\max}$. Because self-attention precedes FFN routing, trajectory queries can attend to observation and XCoT representations even though their FFN updates are produced only by $\mathrm{FFN}_{\mathrm{Control}}$. The Control branch generates $\hat u_t$ with the flow-matching trajectory head and temporally integrates it into the coordinate trajectory $\hat\tau_t$.

\subsection{Stage I: Supervised XCoT and Flow-Matching Training}
\label{sec:sft}

Stage I trains two coupled capabilities: predicting executable XCoT tokens from observations, and generating trajectories conditioned on these tokens. Let $\theta_{\mathrm{reason}}$ denote the parameters of the Reason FFN and XCoT prediction head. The reasoning pathway is supervised by the XCoT sequence produced by the offline data-construction pipeline:
\begin{equation}
\mathcal{L}_{\mathrm{XCoT}}
=
-\mathbb{E}_{(o_t,z_t^*)\sim\mathcal{D}}
\sum_{m=1}^{M_t+1}
\log
\pi_{\theta_{\mathrm{reason}}}(z_{t,m}^*\mid o_t,z_{t,<m}^*),
\qquad
z_{t,M_t+1}^*=\mathrm{EOS}.
\label{eq:xcot_ce}
\end{equation}
Padding tokens are masked out of the loss. During Stage-I training, the trajectory decoder is conditioned on teacher-forced XCoT labels $z_t^*$, while the reasoning pathway is trained to predict the same sequence autoregressively. At inference time, the decoder is conditioned on the predicted sequence $\hat z_t$.

For trajectory generation, we use Conditional Flow Matching (CFM)~\cite{lipman2023flow,tong2024improving}. Let
\[
u_t
=
\{(a_{\mathrm{lon},h},\Delta\psi_h)\}_{h=1}^{H}
\]
be the expert future motion sequence and let $x_1=\mathrm{vec}(u_t)\in\mathbb{R}^{48}$ for $H=24$. Let $x_0\sim\mathcal{N}(0,I_{48})$ be Gaussian noise. For flow time $\alpha\in[0,1]$, define the interpolation $x_\alpha=(1-\alpha)x_0+\alpha x_1$. Given conditioning context $c$ formed from XCoT tokens and trajectory queries, the velocity network $v_\theta$ is trained to predict the target velocity:
\begin{equation}
\mathcal{L}_{\mathrm{FM}}
=
\mathbb{E}_{\alpha,x_0,x_1,c}
\left\|
v_\theta(x_\alpha;\alpha,c)-(x_1-x_0)
\right\|_2^2.
\label{eq:cfm_loss}
\end{equation}
The combined Stage-I objective is:
\begin{equation}
\mathcal{L}_{\mathrm{SFT}}
=
\lambda_{\mathrm{XCoT}}\mathcal{L}_{\mathrm{XCoT}}
+
\lambda_{\mathrm{FM}}\mathcal{L}_{\mathrm{FM}}.
\label{eq:sft_loss}
\end{equation}
Here, $\lambda_{\mathrm{XCoT}}$ and $\lambda_{\mathrm{FM}}$ weight the XCoT prediction and flow-matching objectives, respectively. Unless otherwise stated, $\lambda_{\mathrm{XCoT}}=1$, and $\lambda_{\mathrm{FM}}$ is tuned on a validation split. At inference time, the flow-matching module integrates the learned ODE over flow time $\alpha\in[0,1]$ to obtain the predicted future motion sequence
\[
\hat u_t
=
\{(\hat a_{\mathrm{lon},h},\Delta\hat\psi_h)\}_{h=1}^{H}.
\]
The trajectory head subsequently performs temporal integration over the physical prediction horizon to obtain
\[
\hat\tau_t
=
\mathcal{I}_{\mathrm{traj}}(\hat u_t;s_t).
\]
Flow-time ODE integration over $\alpha$ and temporal integration over the future motion steps $h$ are distinct operations, and neither operation is referred to as an XCoT policy rollout.

\subsection{Stage II: XCoT Policy Optimization}
\label{sec:xcpo}

Stage II optionally refines the XCoT reasoning policy using trajectory-level rewards while keeping the execution stack fixed. During XCPO, the visual encoder, shared multimodal self-attention, Control FFN, and flow-matching trajectory head are frozen as $\theta_{\mathrm{exec}}^{\mathrm{frozen}}$. The deterministic temporal-integration operator has no trainable parameters and is applied after motion generation. Only the Reason FFN and XCoT prediction head, parameterized by $\theta_{\mathrm{reason}}$, are updated.

Given an observation $o$, the old policy samples a group of $G$ XCoT sequences, each of which is decoded by the frozen executor:
\begin{equation}
z_i\sim\pi_{\theta_{\mathrm{old}}}(\cdot\mid o),
\qquad
\tau_i=
\mathcal{F}\!\left(
o,z_i;
\theta_{\mathrm{old}},
\theta_{\mathrm{exec}}^{\mathrm{frozen}}
\right),
\quad i=1,\dots,G,
\label{eq:xcpo_rollout_compact}
\end{equation}
where $\mathcal{F}$ includes reasoning-conditioned representation computation and flow-matching motion generation under the frozen learned modules, followed by parameter-free temporal trajectory integration. Each trajectory receives
\begin{equation}
R_i=\sum_k w_k r_k(\tau_i),
\qquad
A_i=
\frac{R_i-\mu_{\mathbf R}}
{\sigma_{\mathbf R}+\epsilon},
\label{eq:xcpo_reward_compact}
\end{equation}
where $\mu_{\mathbf R}$ and $\sigma_{\mathbf R}$ are the mean and standard deviation within the rollout group. The collected trajectories and rewards are treated as fixed samples during optimization; gradients are propagated only through the autoregressive XCoT log-probabilities.

Following group-relative policy optimization~\cite{shao2024deepseekmath}, we use a clipped sequence-level objective. With
\begin{equation}
\rho_i=
\exp\!\left[
\sum_{m=1}^{M_i+1}
\log
\frac{
\pi_{\theta_{\mathrm{reason}}}(z_{i,m}\mid o,z_{i,<m})
}{
\pi_{\theta_{\mathrm{old}}}(z_{i,m}\mid o,z_{i,<m})
}
\right],
\qquad z_{i,M_i+1}=\mathrm{EOS},
\label{eq:xcpo_ratio_compact}
\end{equation}
the XCPO loss is
\begin{equation}
\begin{aligned}
\mathcal{L}_{\mathrm{XCPO}}
=&-\mathbb{E}_{o}\!\left[
\frac{1}{G}\sum_{i=1}^{G}
\min\!\left(
\rho_iA_i,\,
\operatorname{clip}(\rho_i,1-\varepsilon,1+\varepsilon)A_i
\right)
\right] \\
&+\beta\,
\mathbb{E}_{o}\!\left[
D_{\mathrm{KL}}\!\left(
\pi_{\theta_{\mathrm{reason}}}(\cdot\mid o)
\,\|\,\pi_{\mathrm{ref}}(\cdot\mid o)
\right)
\right],
\end{aligned}
\label{eq:xcpo_loss_compact}
\end{equation}
where $\pi_{\mathrm{ref}}$ is the Stage-I checkpoint. XCPO samples only in the discrete XCoT space. Although the parameters of the execution stack remain frozen, updating the Reason FFN can change the representations that condition trajectory queries in subsequent rollout collection.

\paragraph{Evaluation scope.}
XCPO is an optional policy-refinement extension and is not quantitatively evaluated in the current version. Its intended benefits should be assessed mainly with closed-loop metrics, including route completion, collisions, rule violations, interventions, and comfort, rather than ADE/FDE alone.





\section{Experiments}
\subsection{Experimental Setup}
\label{sec:exp_setup}

\paragraph{Training Data.}
Our model is trained on a large-scale hybrid dataset containing approximately
\textbf{3.6 million unique samples}, as summarized in
\autoref{tab:training_data_distribution}.
The training corpus consists of three components:

\begin{itemize}
    \item \textbf{General SFT (3.1M samples):}
    The main training corpus is collected from large-scale logged driving data
    and automatically labeled using our Reason--Action construction pipeline.

    \item \textbf{Human-Annotated (200k samples):}
    For complex interactive scenarios where automatic annotation can be ambiguous,
    human annotators provide high-quality Reason--Action supervision.

    \item \textbf{Targeted Lane-Change (320k samples):}
    We additionally mine lane-change scenarios to strengthen lateral planning.
    Their XCoT labels are generated using rule-based heuristics to ensure precise
    geometric alignment between semantic actions and logged trajectories.
\end{itemize}

\begin{table}[H]
\centering
\caption{Composition of the training data.}
\label{tab:training_data_distribution}
\small
\setlength{\tabcolsep}{6pt}
\renewcommand{\arraystretch}{1.1}

\begin{tabular}{@{}l l r@{}}
    \toprule
    \textbf{Training Component} &
    \textbf{Supervision Source} &
    \textbf{Samples} \\
    \midrule
    General SFT &
    Automatic Reason--Action labeling &
    3,100,000 \\
    Human-Annotated &
    Human Reason--Action annotation &
    200,000 \\
    Targeted Lane-Change &
    Rule-based XCoT labeling &
    320,000 \\
    \midrule
    \textbf{Total} &
    \textbf{Mixed supervision} &
    \textbf{3,620,000} \\
    \bottomrule
\end{tabular}


\end{table}

\paragraph{Implementation Details.}
We train our model in two stages. In Stage I (SFT), we utilize the full mixed dataset to jointly train the Reason FFN and Control FFN. In Stage II (XCPO), we apply our XCoT Policy Optimization on a subset of safety-critical and complex interaction scenarios to further refine the reasoning policy. The model is trained using the AdamW optimizer with a peak learning rate of $4 \times 10^{-5}$ for Stage I and $1 \times 10^{-4}$ for Stage II.

\subsection{Evaluation Protocol}
We evaluate XCoT-VLA from three complementary perspectives: open-loop planning accuracy, reasoning-interface efficiency, and optimization stability. The open-loop evaluation uses a general-distribution set and a lane-change set. We additionally profile latency under a 12\,Hz planning budget, analyze the stability of XCoT fine-tuning, and present qualitative trajectory cases covering dense traffic, navigation-guided preemptive lane changes, traffic-light compliance, and efficiency-driven lane changes.

For open-loop planning, we report directional Average Displacement Error (ADE) and Final Displacement Error (FDE) in an ego-centric route-aligned coordinate frame. Our default evaluation horizon is \textbf{6\,s}, corresponding to $H=24$ future steps. Therefore, unless explicitly marked as \emph{ADE-2s}, all ADE and FDE values in this section are computed over the full 6\,s horizon and are denoted as \emph{ADE-6s} and \emph{FDE-6s}, respectively. The additional \emph{ADE-2s} metric averages displacement error over the first 2\,s ($H_{2s}=8$ steps) and is reported only in the training-stability analysis.

Let
$\hat{\tau}_i=\{(\hat{x}_{i,h},\hat{y}_{i,h})\}_{h=1}^{H}$ and
$\tau_i=\{(x_{i,h},y_{i,h})\}_{h=1}^{H}$ denote the predicted and ground-truth trajectories for sample $i$, respectively, where $x$ and $y$ are the longitudinal and lateral coordinates. For a set of $N$ samples, the full-horizon directional ADE metrics are
\begin{equation}
\label{eq:directional_ade}
\begin{aligned}
\mathrm{ADE\mbox{-}6s\mbox{-}Long}
&=
\frac{1}{NH}
\sum_{i=1}^{N}\sum_{h=1}^{H}
\left|\hat{x}_{i,h}-x_{i,h}\right|,\\
\mathrm{ADE\mbox{-}6s\mbox{-}Lat}
&=
\frac{1}{NH}
\sum_{i=1}^{N}\sum_{h=1}^{H}
\left|\hat{y}_{i,h}-y_{i,h}\right|.
\end{aligned}
\end{equation}
The corresponding terminal errors at 6\,s are
\begin{equation}
\label{eq:directional_fde}
\begin{aligned}
\mathrm{FDE\mbox{-}6s\mbox{-}Long}
&=
\frac{1}{N}
\sum_{i=1}^{N}
\left|\hat{x}_{i,H}-x_{i,H}\right|,\\
\mathrm{FDE\mbox{-}6s\mbox{-}Lat}
&=
\frac{1}{N}
\sum_{i=1}^{N}
\left|\hat{y}_{i,H}-y_{i,H}\right|.
\end{aligned}
\end{equation}
For the short-horizon diagnostic, we additionally report
\begin{equation}
\label{eq:directional_ade_2s}
\begin{aligned}
\mathrm{ADE\mbox{-}2s\mbox{-}Long}
&=
\frac{1}{N H_{2s}}
\sum_{i=1}^{N}\sum_{h=1}^{H_{2s}}
\left|\hat{x}_{i,h}-x_{i,h}\right|,\\
\mathrm{ADE\mbox{-}2s\mbox{-}Lat}
&=
\frac{1}{N H_{2s}}
\sum_{i=1}^{N}\sum_{h=1}^{H_{2s}}
\left|\hat{y}_{i,h}-y_{i,h}\right|.
\end{aligned}
\end{equation}
The predicted coordinates used by these metrics are obtained from the temporally integrated output of the trajectory head, $\hat\tau_i=\mathcal{I}_{\mathrm{traj}}(\hat u_i;s_i)$. All displacement metrics are reported in meters, and lower values indicate better trajectory accuracy. Unlike conventional Euclidean ADE/FDE, these metrics separately measure longitudinal and lateral displacement errors. Relative improvement is computed as $(\mathrm{baseline}-\mathrm{method})/\mathrm{baseline}$.

We compare four controlled variants: \emph{Trajectory only}, \emph{Verbose CoT}, \emph{Latent tokens}, and \emph{XCoT-VLA}. Here, \emph{Trajectory only} denotes our standard trajectory-supervised SFT baseline. They share the same visual backbone, trajectory decoder, prediction horizon, and open-loop evaluation protocol unless stated otherwise. \autoref{tab:variant_definitions} summarizes their reasoning interfaces and supervision.

\begin{table*}[!htbp]
    \centering
    \captionsetup{font=small,skip=4pt}
    \caption{Controlled model variants used in the experiments.}
    \label{tab:variant_definitions}
    \footnotesize
    \setlength{\tabcolsep}{6pt}
    \renewcommand{\arraystretch}{1.18}
    \begin{tabularx}{\textwidth}{@{}>{\RaggedRight\arraybackslash}p{0.17\textwidth}
        >{\RaggedRight\arraybackslash}p{0.18\textwidth}
        >{\RaggedRight\arraybackslash}p{0.24\textwidth}
        >{\RaggedRight\arraybackslash}X@{}}
        \toprule
        \textbf{Variant} & \textbf{Reasoning interface} & \textbf{Supervision} &
        \textbf{Decoder / training} \\
        \midrule
        Trajectory only(SFT) & None & Logged trajectories only &
        Same trajectory decoder without explicit reasoning tokens. \\
        Verbose CoT & Natural-language-style rationale & Verbose semantic supervision &
        Same decoder with a long, descriptive reasoning interface. \\
        Latent tokens & Compact semantic tokens &
        Compressed labels without joint Reason--Action assignment &
        Same decoder and training data, but without the causal-reason and maneuver semantics used by XCoT. \\
        \rowcolor{gray!12}
        \textbf{XCoT-VLA} & Executable XCoT sequence &
        Reason--Action supervision from trajectory and scene context &
        Shared self-attention with separate Reason and Control FFN branches. \\
        \bottomrule
    \end{tabularx}
\end{table*}

\subsection{Main Results: Open-Loop Planning}

\autoref{tab:openloop_main} reports the main open-loop results. XCoT-VLA achieves the best performance on all reported metrics for both evaluation sets. On the general set, ADE-6s-Long decreases from 1.6452 to 1.3233 and FDE-6s-Long from 4.3541 to 3.0887 relative to Trajectory only. The gains are larger on the lane-change set, where ADE-6s-Lat decreases from 0.5941 to 0.3091 and FDE-6s-Lat from 1.6160 to 0.6484. Verbose CoT does not consistently improve lateral accuracy, while Latent tokens help but remain behind the full Reason--Action supervision of XCoT-VLA. These results indicate that executable reasoning is particularly effective for route-conditioned lateral decisions.

\begin{table*}[!htbp]
    \centering
    \captionsetup{font=small,skip=4pt}
    \caption{Open-loop planning results on the general-distribution and lane-change evaluation sets. ADE-6s and FDE-6s are computed over the full 6\,s prediction horizon. Lower is better.}
    \label{tab:openloop_main}
    \footnotesize
    \setlength{\tabcolsep}{6pt}
    \renewcommand{\arraystretch}{1.18}
    \begin{tabularx}{\textwidth}{@{}>{\centering\arraybackslash}p{0.13\textwidth}
        >{\RaggedRight\arraybackslash}p{0.24\textwidth}
        *{4}{>{\centering\arraybackslash}X}@{}}
        \toprule
        \textbf{Set} & \textbf{Method} & \textbf{ADE-6s-Lat} & \textbf{ADE-6s-Long} &
        \textbf{FDE-6s-Lat} & \textbf{FDE-6s-Long} \\
        \midrule
        \multirow[c]{4}{*}{General}
        & Trajectory-only(SFT) & 0.2609 & 1.6452 & 0.7352 & 4.3541 \\
        & Verbose CoT & 0.2798 & 1.4382 & 0.7822 & 3.5045 \\
        & Latent tokens & 0.2511 & 1.3738 & 0.6804 & 3.2861 \\
        \rowcolor{gray!12}
        & \textbf{XCoT-VLA} & \textbf{0.2162} & \textbf{1.3233} &
        \textbf{0.5765} & \textbf{3.0887} \\
        \midrule
        \multirow[c]{4}{*}{Lane-change}
        & Trajectory-only(SFT) & 0.5941 & 1.8221 & 1.6160 & 4.8399 \\
        & Verbose CoT & 0.3589 & 1.6217 & 0.7692 & 3.8088 \\
        & Latent tokens & 0.3347 & 1.5453 & 0.7102 & 3.5303 \\
        \rowcolor{gray!12}
        & \textbf{XCoT-VLA} & \textbf{0.3091} & \textbf{1.4552} &
        \textbf{0.6484} & \textbf{3.2258} \\
        \bottomrule
    \end{tabularx}
\end{table*}

\FloatBarrier

\subsection{Ablation: XCoT and Navigation Supervision}

\autoref{tab:navigation_ablation} separates the contribution of XCoT supervision from that of targeted navigation data. XCoT alone reduces ADE-6s-Lat by 18.3\% and FDE-6s-Lat by 25.1\% relative to the trajectory-only SFT baseline. Adding navigation supervision further reduces ADE-6s-Lat to 0.4518 and FDE-6s-Lat to 1.1122. The smaller longitudinal changes are consistent with this ablation primarily affecting route-conditioned lateral intent.

\begin{table*}[!htbp]
    \centering
    \captionsetup{font=small,skip=4pt}
    \caption{Ablation on XCoT reasoning and targeted navigation data on a controlled lane-change diagnostic split. ADE-6s and FDE-6s are computed over the full 6\,s prediction horizon. Lower is better. This split should not be compared numerically with \autoref{tab:openloop_main}.}
    \label{tab:navigation_ablation}
    \footnotesize
    \setlength{\tabcolsep}{6pt}
    \renewcommand{\arraystretch}{1.18}
    \begin{tabularx}{\textwidth}{@{}>{\RaggedRight\arraybackslash}p{0.22\textwidth}
        *{6}{>{\centering\arraybackslash}X}@{}}
        \toprule
        \textbf{Method} & \textbf{XCoT} & \textbf{Nav.} & \textbf{ADE-6s-Lat} &
        \textbf{ADE-Long} & \textbf{FDE-6s-Lat} & \textbf{FDE-6s-Long} \\
        \midrule
        Trajectory-only(SFT) & -- & -- & 0.5941 & 1.8221 & 1.6160 & 4.8399 \\
        XCoT only & \checkmark & $\times$ & 0.4856 & 1.8418 & 1.2106 & 4.8458 \\
        \rowcolor{gray!12}
        \textbf{XCoT + Nav.} & \checkmark & \checkmark &
        \textbf{0.4518} & \textbf{1.7977} & \textbf{1.1122} & \textbf{4.7230} \\
        \bottomrule
    \end{tabularx}
\end{table*}

\subsection{Efficiency of Executable XCoT}

XCoT replaces a verbose natural-language rationale with a compact 2--6-token executable interface. In typical driving reasoning, a conventional CoT rationale may span dozens of autoregressive tokens to describe the scene, explain causal factors, and articulate the intended maneuver; in our setting, 40--80 tokens is a representative range for such verbose CoT outputs. Here, \emph{time to first token} (TTFT) denotes the latency from submitting the multimodal input to generating the first reasoning token, including input-context prefill and the initial decoding step. We measure an inter-token latency of 3.25\,ms on an H100 and estimate the total reasoning-interface latency as $\mathrm{TTFT}+3.25M$, where $M$ is the number of generated reasoning tokens. \autoref{tab:efficiency_latency} reports the conservative worst case using $M=80$ for Verbose CoT and $M=6$ for XCoT-VLA. XCoT-VLA remains below the 83.3\,ms budget for all evaluated input lengths, whereas Verbose CoT exceeds this budget throughout. This analysis covers only the reasoning interface and excludes perception preprocessing, trajectory postprocessing, and full-stack scheduling overhead.

\begin{table*}[!htbp]
    \centering
    \captionsetup{font=small,skip=4pt}
    \caption{Worst-case reasoning-interface latency on an H100. Results use each method's maximum output length; the 12\,Hz budget is 83.3\,ms.}
    \label{tab:efficiency_latency}
    \footnotesize
    \setlength{\tabcolsep}{6pt}
    \renewcommand{\arraystretch}{1.18}
    \begin{tabularx}{\textwidth}{@{}>{\RaggedRight\arraybackslash}m{0.19\textwidth}
        >{\centering\arraybackslash}m{0.12\textwidth}
        *{4}{>{\centering\arraybackslash}X}
        >{\centering\arraybackslash}m{0.11\textwidth}@{}}
        \toprule
        \multirow[c]{2}{*}{\textbf{Method}} &
        \multirow[c]{2}{*}{\shortstack[c]{\textbf{Output}\\\textbf{tokens}}} &
        \multicolumn{4}{c}{\textbf{Worst-case latency (ms)}} &
        \multirow[c]{2}{*}{\shortstack[c]{\textbf{Within}\\\textbf{12\,Hz budget}}} \\
        \cmidrule(lr){3-6}
        & & \textbf{2K} & \textbf{3.5K} & \textbf{5K} & \textbf{6K} & \\
        \midrule
        Verbose CoT & 40--80 & 279.1 & 287.8 & 298.2 & 306.8 & $\times$ \\
        \rowcolor{gray!12}
        \textbf{XCoT-VLA} & \textbf{2--6} & \textbf{38.6} & \textbf{47.3} &
        \textbf{57.7} & \textbf{66.3} & \checkmark \\
        \bottomrule
    \end{tabularx}
    \vspace{1mm}

\end{table*}

\FloatBarrier



\subsection{Training Stability}

\autoref{tab:forgetting} compares the trajectory-only SFT baseline with joint and decoupled XCoT fine-tuning (FT). Joint XCoT FT causes a pronounced longitudinal degradation: ADE-6s-Long increases from 1.2997 to 1.6005 and ADE-2s-Long from 0.2774 to 1.1684. Decoupled XCoT FT preserves short-horizon longitudinal accuracy and improves ADE-6s-Lat from 0.2302 to 0.1872 and FDE-6s-Lat from 0.5710 to 0.4690. Its remaining weakness is long-horizon longitudinal accuracy, where ADE-6s-Long and FDE-6s-Long remain above the SFT baseline. These results support separating the Reason and Control FFN branches during XCoT fine-tuning.

\begin{table*}[!htbp]
    \centering
    \captionsetup{font=small,skip=4pt}
    \caption{Training stability under joint and decoupled XCoT fine-tuning. ADE-6s/FDE-6s use the full 6\,s horizon, while ADE-2s uses the first 2\,s. Lower is better.}
    \label{tab:forgetting}
    \footnotesize
    \setlength{\tabcolsep}{5pt}
    \renewcommand{\arraystretch}{1.18}
    \begin{tabularx}{\textwidth}{@{}>{\RaggedRight\arraybackslash}p{0.23\textwidth}
        *{6}{>{\centering\arraybackslash}X}@{}}
        \toprule
        \textbf{Training strategy} & \textbf{ADE-6s-Lat} & \textbf{ADE-6s-Long} &
        \textbf{ADE-2s-Lat} & \textbf{ADE-2s-Long} &
        \textbf{FDE-6s-Lat} & \textbf{FDE-6s-Long} \\
        \midrule
        Trajectory-only(SFT) & 0.2302 & \textbf{1.2997} & 0.0360 &
        0.2774 & 0.5710 & \textbf{3.2521} \\
        Joint XCoT FT & 0.2774 & 1.6005 & 0.0355 &
        1.1684 & 0.7313 & 3.2854 \\
        \rowcolor{gray!12}
        \textbf{\mbox{Decoupled XCoT FT}} & \textbf{0.1872} & 1.3330 &
        \textbf{0.0313} & \textbf{0.2766} & \textbf{0.4690} & 3.3864 \\
        \bottomrule
    \end{tabularx}
\end{table*}

\begin{figure}[!b]
\centering
\captionsetup{font=small,skip=3pt}
\safeincludegraphics[
    width=\textwidth,
    height=0.70\textheight,
    keepaspectratio
]{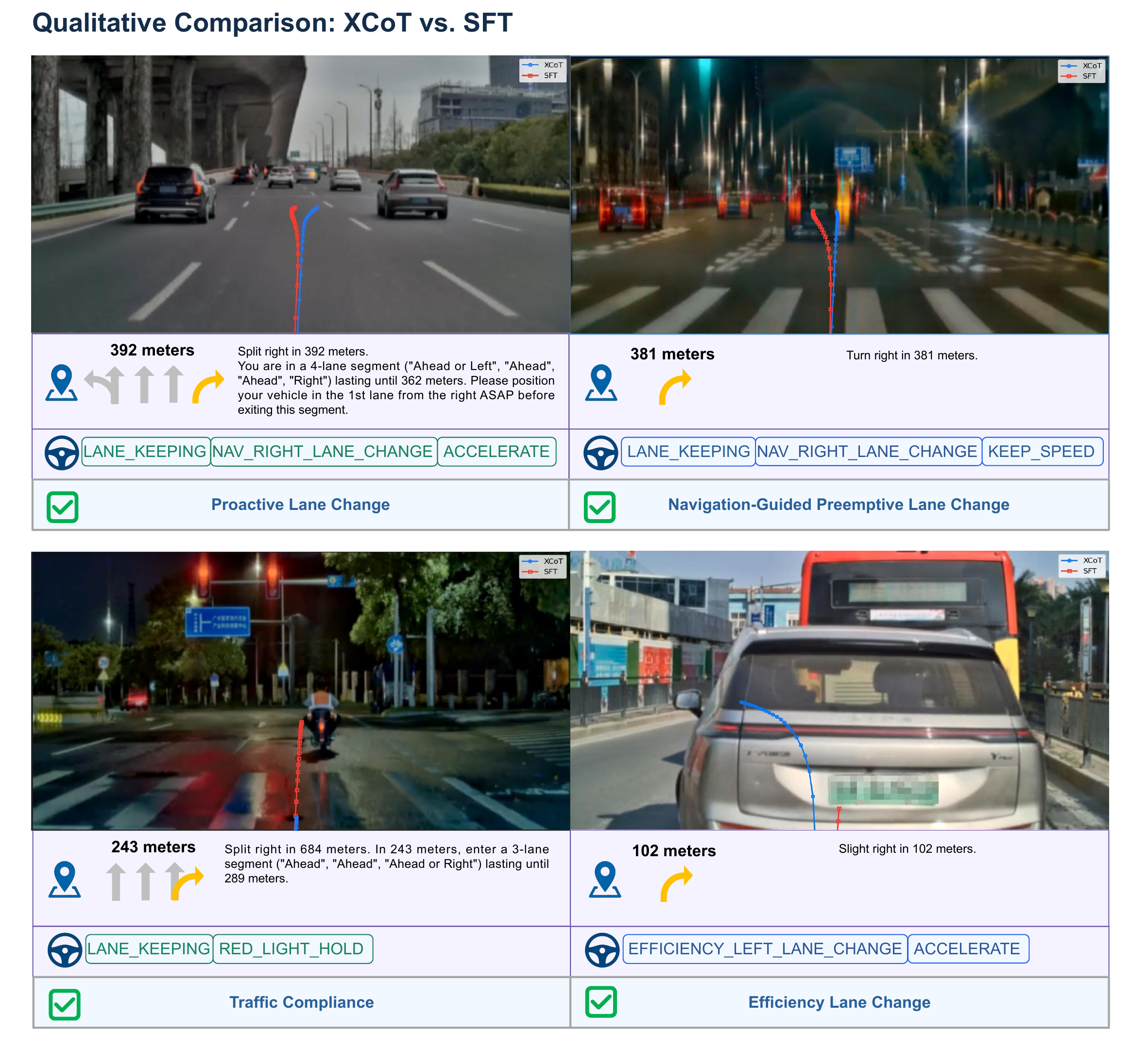}
\caption{Qualitative comparison between XCoT (blue) and the trajectory-only SFT baseline (red). Compared with SFT, XCoT consistently exhibits more proactive driving behaviors, including earlier lane changes in dense traffic and navigation-guided scenarios, proactive compliance with traffic-light constraints, and efficiency-driven lane changes with traffic-adaptive acceleration, leading to smoother and more efficient trajectory planning.}
\label{fig:qualitative_xcot}
\vspace{-2mm}
\end{figure}

\subsection{Qualitative Analysis}
\label{sec:qualitative_xcot}
\autoref{fig:qualitative_xcot} presents four representative open-loop cases covering dense-traffic lane changes, navigation-guided preemptive lane changes, traffic-light compliance, and efficiency-driven lane changes. Compared with the SFT baseline, XCoT exhibits more proactive and context-aware behaviors, including earlier lane changes, proactive responses to traffic-light constraints, and traffic-adaptive acceleration for efficient maneuvering. These cases illustrate how executable XCoT tokens translate route, lateral, longitudinal, and traffic-rule semantics into improved trajectory planning.


\section{Conclusion}

We presented XCoT-VLA, which replaces verbose natural-language CoT with 2--6 executable semantic-action tokens learned from Reason--Action supervision. All non-trajectory tokens are processed by the Reason FFN, while trajectory queries use the Control FFN to generate continuous motion after shared multimodal self-attention. XCoT-VLA improves open-loop planning accuracy, particularly in lane-change scenarios, while substantially reducing autoregressive reasoning cost.

These results support compact, action-facing reasoning as a practical interface between semantic understanding and trajectory generation. Nevertheless, XCoT currently has limited capability for interactive negotiation with surrounding agents, remains sensitive to traffic-rule perception errors, and shows smaller improvements in longitudinal comfort. Future work will strengthen interaction- and uncertainty-aware reasoning, extend XCoT to richer rule-governed scenarios, and quantitatively evaluate XCPO using closed-loop safety, route-completion, and comfort metrics.
\section{Contributors}
We sincerely thank every member of the team for their dedication and valuable contributions. This work reflects our ongoing efforts in advancing reasoning-driven Vision-Language-Action models for autonomous driving. XCoT-VLA introduces a compact and executable reasoning interface that directly connects multimodal scene understanding with trajectory generation.

\textbf{Advisors}: Hang Zhang, Honggou Yang, Xianming Liu

\textbf{Project Lead}: Qiman Wu

\textbf{Contributors}:
Yue He\textsuperscript{*},
Shaman Tang\textsuperscript{*},
Dong Xing\textsuperscript{*},
Hui Xue\textsuperscript{*},
Lvjie Chen,
Menglin Li,
Hanlin Chen,
Hua Zhou,
Yizhao Wang,
Anhua Liu,
Yuheng Zhang,
Mingyuan Wang,
Rui Xin,
Jiahui Hu,
Da Zhu,
Yuhua Wei,
Pengfei Diao,
Shuang Su,
Minghao Li,
Haojie Yang,
Siqi Liu,
Kai Yang,
ZhenKang Wu,
Weichao Huang.

\textsuperscript{*}Core contribution.

{
\begingroup
\sloppy
\small
\bibliographystyle{turing_unsrtnat}
\bibliography{main}
\endgroup
}

\end{document}